\documentclass[conference]{IEEEtran}
\IEEEoverridecommandlockouts
\usepackage{cite}
\usepackage{amsmath,amssymb,amsfonts}
\usepackage{algorithmic}
\usepackage{graphicx}
\usepackage{textcomp}
\usepackage{xcolor}
\newcommand{\argmax}{\operatornamewithlimits{argmax}}
\usepackage{multirow}
\usepackage{makecell}

\usepackage{xcolor}
\definecolor{forestgreen}{rgb}{0.0, 0.27, 0.13}
\definecolor{falured}{rgb}{0.5, 0.09, 0.09}
\definecolor{desert}{rgb}{0.69,0.4,0.0}
\definecolor{ao}{rgb}{0.0, 0.5, 0.0}
\def\BibTeX{{\rm B\kern-.05em{\sc i\kern-.025em b}\kern-.08em
    T\kern-.1667em\lower.7ex\hbox{E}\kern-.125emX}}
\begin{document}

\title{SE-DAE: Style-Enhanced Denoising Auto-Encoder for Unsupervised Text Style Transfer\\
}

% \and
% \and

\author{
\IEEEauthorblockN{Jicheng Li, Yang Feng*\thanks{*The corresponding author is Yang Feng.}, Jiao Ou}
\IEEEauthorblockA{\textit{Key Laboratory of Intelligent Information Processing}, \\ \textit{Institute of Computing Technology, Chinese Academy of Sciences (ICT/CAS)}, Beijing, China}
\IEEEauthorblockA{\textit{University of Chinese Academy of Sciences}, Beijing, China}
lijicheng@ict.ac.cn,fengyang@ict.ac.cn,oujiao17b@ict.ac.cn
}

\maketitle

\begin{abstract}
Text style transfer aims to change the style of sentences while preserving the semantic meanings. Due to the lack of parallel data, the Denoising Auto-Encoder (DAE) is widely used in this task to model distributions of different sentence styles. However, because of the conflict between the target of the conventional denoising procedure and the target of style transfer task, the vanilla DAE can not produce satisfying enough results. To improve the transferability of the model, most of the existing works combine DAE with various complicated unsupervised networks, which makes the whole system become over-complex. 
In this work, we design a novel DAE model named Style-Enhanced DAE (SE-DAE), which is specifically designed for the text style transfer task.
Compared with previous complicated style-transfer models, our model do not consist of any complicated unsupervised networks, but only relies on the high-quality pseudo-parallel data generated by a novel data refinement mechanism. Moreover, to alleviate the conflict between the targets of the conventional denoising procedure and the style transfer task, we propose another novel style denoising mechanism, which is more compatible with the target of the style transfer task. We validate the effectiveness of our model on two style benchmark datasets. Both automatic evaluation and human evaluation show that our proposed model is highly competitive compared with previous strong the state of the art (SOTA) approaches and greatly outperforms the vanilla DAE.
\end{abstract}

% \begin{IEEEkeywords}
% text style transfer, denoising auto-encoder, refinement
% \end{IEEEkeywords}

\section{Introduction}
Text style transfer is an intractable challenge, focusing more on how to change the style-attribute of sentences while preserving their semantics. 
Due to the lack of paired sentences, the Denoising Auto-Encoder (DAE) has been used widely in existing style-transfer models to model distributions of different styles in unsupervised manners. However, the target of the conventional denoising process that requires models to recover all the information in the sentences conflicts with the target of the style transfer task that only requires models to preserve the semantic information but transfer the style information. As a consequence, the vanilla DAE will regard rare style words as content words and preserve them in final transferred results. Thus, the vanilla DAE can not produce satisfying enough results in this task.

To enhance the transferability of DAE, most of the proposed transfer models combine DAE with different unsupervised neural networks, especially Reinforcement Learning (RL) \cite{cycle-rl,wu-rl,luo19a,luo19b,gong-rl} and Generative Adversarial Networks (GANs) \cite{hu-gan,fu-gan,Lajanugen-gan,yang-gan,lai-gan,style-gan,tikhonov-etal-2019-style}.However, since these proposed style-transfer models (e.g., RL-based and GAN-based models) are composed of complicated neural networks, it is difficult to converge \cite{lample2019,ub-gan}, which also requires over-complex and laborious training procedures. Furthermore, the complex training procedures also make it intricate to adapt these models to new style datasets. To simplify the training procedure, Lample et al.\cite{lample2019} proposes training DAE-based models on the pseudo-parallel data, rather than using it in combination with any unsupervised neural network. Nevertheless, the pseudo-parallel data generated by back-translation is plagued by severe data-quality problem \cite{luo19a}, which adversely affects the performance of the model.

In general, the conventional DAE-based models have the following two conflicts: (1) the conflict between the goal of the conventional denoising process and the goal of the style transfer task; (2) the conflict between simplicity and performance of the model. To resolve the above conflicts, we propose a novel DAE model with two task-oriented mechanisms, namely the the Style-Enhanced Denoising Auto-Encoder (SE-DAE). We deal with the first conflict through a novel style denoising mechanism, that is, we randomly ``contaminate'' the input sentences with style words during the corruption process, and require the model to remove the contaminants and recover the original sentence during the reconstruction process. In this way, the goal of the style denoising mechanisms is compatible with the goal of the text style transfer task. Next, we propose a novel data refinement mechanism to select the best inference to construct a high-quality pseudo-parallel dataset, which solves the data-quality problem. By using the high-quality pseudo-parallel data for training on the DAE model, we enhance the transferability of the model without combining any complex unsupervised networks. Finally, we evaluate the performance of our model on two benchmark datasets: \textsc{Yelp} \cite{li-2018-delete} for the sentiment transfer task and \textsc{Gyafc} \cite{rao-tetreault-2018-dear} for the formality transfer task. Both automatic and human evaluation show that our model is highly competitive compared with previous strong SOTAs and greatly outperforms the most commonly used vanilla DAE.

\section{Background}

\subsection{Denoising Auto-Encoder (DAE)}
DAE \cite{dae} is an unsupervised neural network that aims to learn robust representations by corrupting input pattern partially, which can be utilized to initialize deep neural architectures. In general, the training process of the DAE model is composed of corruption and reconstruction.

For an input $\bf s$ in dataset $\mathcal{D}$, DAE model first corrupt $\bf s$ to a noise version $\bf c(\bf s)$ and then encoder $\bf{E}(c(\bf s);\theta_{E})$ maps $\bf c(\bf s)$ into a latent representation $\bf z$:

\begin{equation}
    \bf{z} = E(c(\bf s);\theta_{E})
\end{equation}
Then, decoder $D(\bf z;\theta_{D})$ reconstructs the original input $\bf s$ condition on $\bf z$:
\begin{equation}
    \bf{\hat s} = D(\bf z;\theta_{D})
\end{equation}
Finally, the objective of the DAE is to minimize the following loss:
\begin{equation}
    \mathcal{L}_{\bf{DAE}}=\mathbb{E}_{\bf{s} \sim \mathcal{D},\bf{\hat{ s}} \sim D(E(c(\bf{s})))}[\Delta(\bf{s},\bf{\hat{s}})] 
\end{equation}

where $\Delta(\bf{s},\bf{\hat s})$ is a measure of the discrepancy between $\bf{s}$ and $\bf{\hat s}$.

\begin{figure}[!t]
\centering
\centerline{\includegraphics[width=.9\columnwidth]{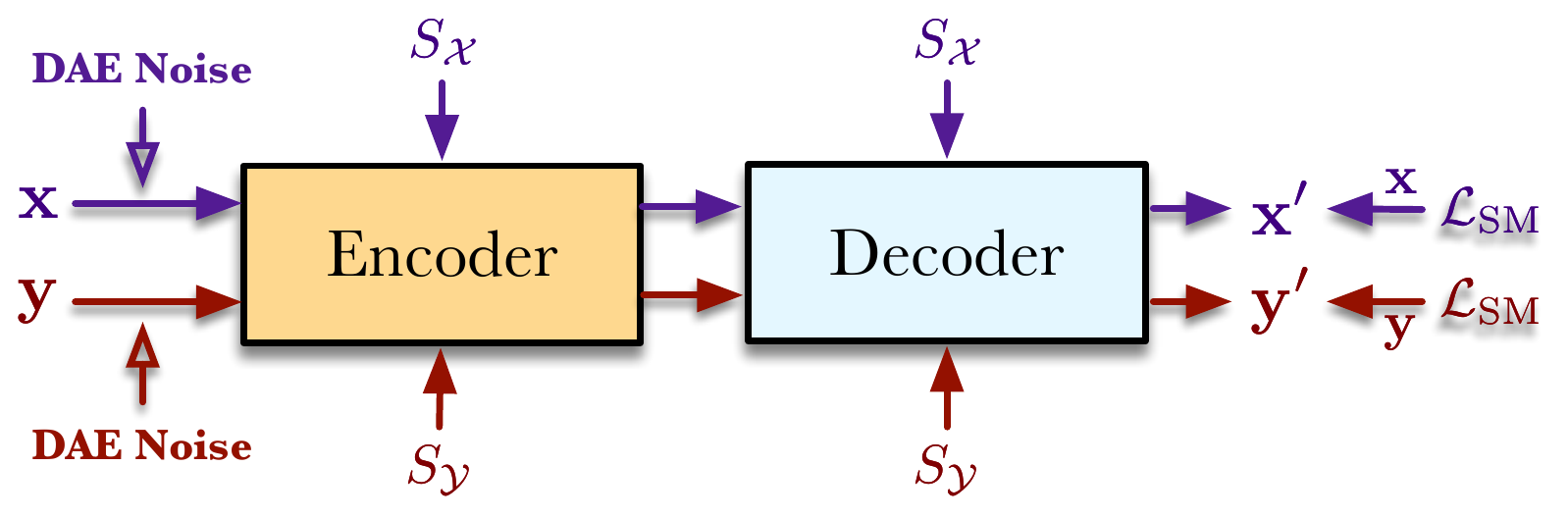}}
\caption{Style modelling process of style $\mathcal{X}$ and $\mathcal{Y}$. ``DAE Noise'' represents \textit{dropping} and \textit{shuffling} words. ``$\bf{S_{\mathcal{X}}}$'' and ``$\bf{S_{\mathcal{Y}}}$'' represent embedding of style $\mathcal{X}$ and $\mathcal{Y}$ respectively.} 
\vspace{-1em}
\label{fig:1} 
\end{figure}

\subsection{Text Style Transfer Task}
Given a set of corpus with different styles $\{\mathcal{D}_{i}\}^{K}_{i=1}$ where each corpus only contains sentences with the same style, $\mathcal{D}_{\mathcal{X}}=\{\bf x_{\bf i}\}^{n}_{i=1}$ is the corpus of source style $\mathcal{X}$ and $\mathcal{D}_{\mathcal{Y}}=\{\bf y_{\bf i}\}^{m}_{i=1}$ is the corpus of target style $\mathcal{Y}$ ($\mathcal{Y} \neq \mathcal{X}$). For each sentences $\bf x$ in $\mathcal{D}_{\mathcal{X}}$, the text style transfer task aims to rewrite $\bf x$ into $\bf{x}_{_\mathcal{Y}}$ with the attributes of desired style $\mathcal{Y}$ and the semantics meaning of $\bf x$.

\section{Basic Structure}

\subsection{Style Modeling} \label{sec:lm}
Like Lample et al.\cite{lample-etal-2018-phrase} modeling different languages via denoise auto-encoding, DAE can be also used to model distributions of different styles as shown in Figure \ref{fig:1}. 
%for input sentences $\bf x$ of style $\mathcal{X}$ and $\bf y$ of style $\mathcal{Y}$, 
Likewise, we corrupt the input sentences and train the DAE to use the corrupted sentence as input to reconstruct the original sentences. More specifically, when corrupting sentences, we consider two types of noises in our work when corrupting sentences: \textit{dropping} and \textit{shuffling} input tokens.
Besides, to unify latent spaces of different styles, we share encoders and decoders between styles and utilize style embedding (for instance, $\bf S_{\mathcal{X}}$ and $\bf S_{\mathcal{Y}}$ for style $\mathcal{X}$ and $\mathcal{Y}$ respectively) to indicate different styles during training and inference. At last, the token-level cross-entropy is used to measure the discrepancy between the original sentences and the reconstructed sentences. 

Overall, the loss of style modelling can be expressed as:
\begin{equation}
    \mathcal{L}_{\bf{SM}}=\log{\bf P(\bf x | c(\bf x), S_{\mathcal{X}})}
  +\log{\bf P(\bf y | c(\bf y),S_{\mathcal{Y}})} \label{eq:lm}
\end{equation}

\subsection{Iterative Back Translation}
The style modeling process can only help the DAE to model the distribution of each style, but not the way of transferring sentences from one style to another style. Most of the previous works combine the DAE with a complicated unsupervised models like RL and GANs, to learn how to transfer styles of sentences. Although this combinations can create a powerful style-transfer model, it makes the training procedure of the entire model becomes over-complex. Instead of seeking helps from complicated unsupervised models, we train the DAE to transfer between styles by iterative back-translation \cite{lample2019} to keep the simplicity of DAE to the greatest extent possible. Every epoch, we use the DAE at the last epoch to back-translate the unpaired data into the pseudo-parallel data and continue to train it upon the recently generate pseudo-parallel data. Specifically, we utilize the DAE model at epoch $\bf t$, named ${\bf M}_{\mathcal{X} \Leftrightarrow \mathcal{Y}}^{\bf t}$, to back-translate $\bf x$ and $\bf y$ to $\bf{\hat{y}}$ of style $\mathcal{Y}$ and $\bf{\hat{x}}$ of $\mathcal{X}$ respectively and pair the corresponding sentences to form the pseudo-parallel data $[\bf{\hat{y}},\bf x]$ and $[\bf{\hat{x}},\bf y]$. Then, we train ${\bf M}_{\mathcal{X} \Leftrightarrow \mathcal{Y}}^{\bf t}$ on the recently generated pseudo-parallel data and obtain the DAE model at epoch $\bf t+1$, named ${\bf M}_{\mathcal{X} \Leftrightarrow \mathcal{Y}}^{\bf t+1}$. In general, the loss of iterative back translation in each epoch can be represented as follow:
\begin{equation}
    \mathcal{L}_{\bf{BT}}=\log{\bf P(\bf{y}  | \bf{\hat{x}})}
  +\log{\bf P(\bf x | \bf{\hat{y}})} \label{eq:lm}
\end{equation}

\begin{figure*}
\centering
\includegraphics[width=1.5\columnwidth]{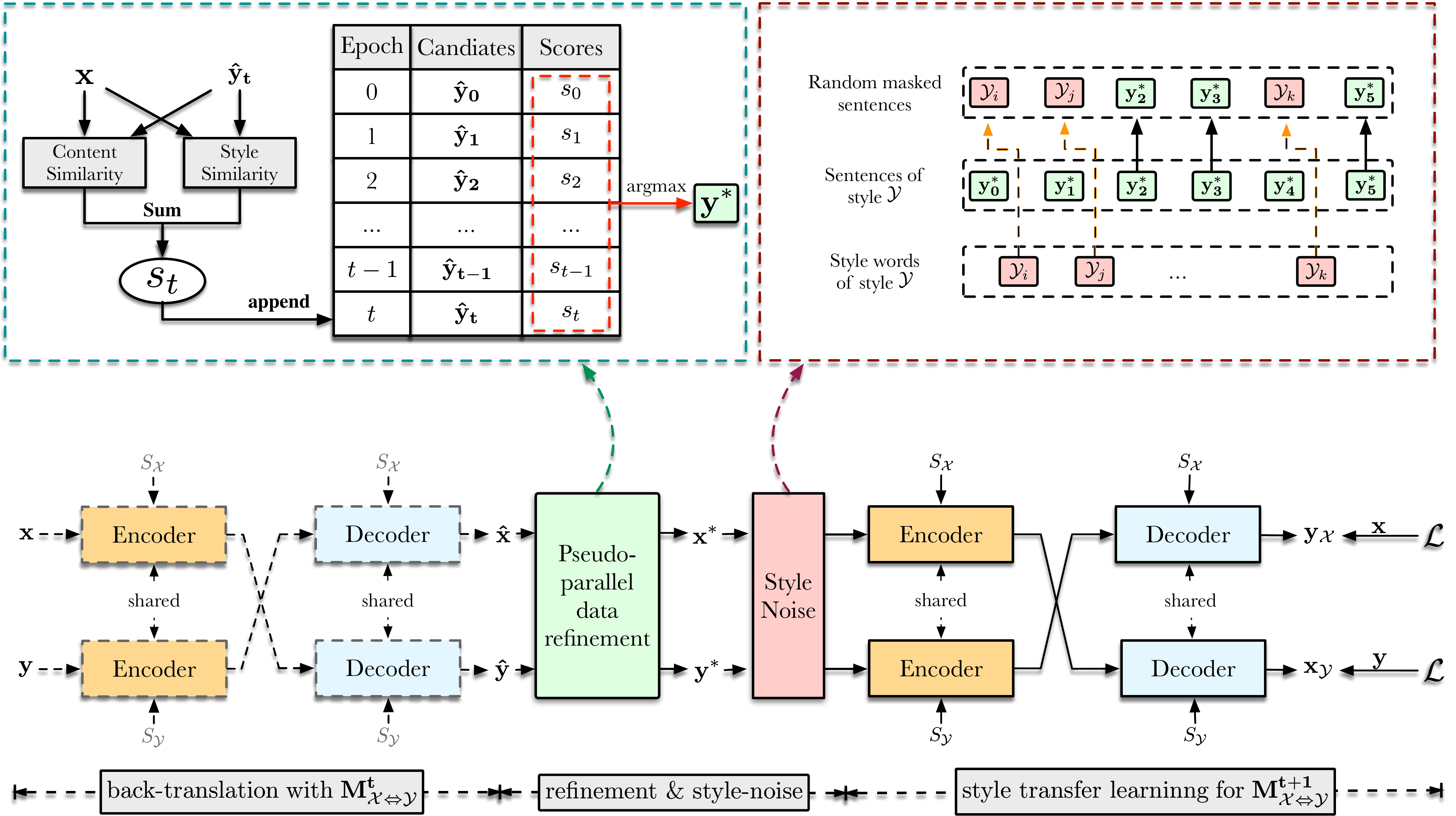}
\caption{The training process of our model. ${\bf M}_{\mathcal{X} \Leftrightarrow \mathcal{Y}}^{\bf t}$ represents the style-transfer model at epoch $\bf t$, which is utilized to generate pseudo-parallel data to train ${\bf M}_{\mathcal{X} \Leftrightarrow \mathcal{Y}}^{\bf t+1}$. } 
\vspace{-1em}
\label{fig:2} 
\end{figure*}

\section{Our Method} \label{sec:stl}
In this section, we will introduce our style-enhanced DAE model with two task-oriented modules, as shown in the Figure \ref{fig:2}. 

\subsection{Pseudo-parallel Data Reﬁnement} \label{sec:g&r}
Unfortunately, although the pseudo-parallel data can help DAE to learn how to transfer styles of sentences, it suffers from the data-quality problem \cite{luo19a} which limits the performance of DAE in this task. 
In this work, we propose an new mechanism to refine the pseudo-parallel data by finding the best inference in the history of all training epochs.

In epoch $\bf t$, for each sentence $\bf x$ of style $\mathcal{X}$, we append the back-translated sentence $\bf{\hat{y}^{t}}$ of style $\mathcal{Y}$ into the candidates set of $\bf x$ : $\bf \mathbb{C}_{\bf x}=\{\bf \hat{y}^{0},\hat{y}^{1},\ldots,\hat{y}^{t} \}$. Since we prefer to choose $\bf \hat{y}$ which preserves the most semantics of $\bf x$ and has the highest intensity towards $\mathcal{Y}$. Thus, we design a scoring function $\bf{CS}$, which estimates the quality of each $\bf{\hat{y}}$ in $\bf \mathbb{C}_{\bf x}$ from two aspects: \textbf{the semantic similarity between $\bf x$ and $\bf \hat{y}$ } and \textbf{the style intensity towards $\mathcal{Y}$}. Similarly, we adopt same operations to the sentence $\bf \hat{x}$ in the candidate set $\bf \mathbb{C}_{\bf y}$.

\subsubsection{Semantic Similarity} 
In this work, we adopt the \textit{Word Mover Distance} (WMD) \cite{wmd} to estimate the semantic similarity between $\bf{\hat{y}}$ and $\bf x$:
\begin{equation}
    \bf {Con}(\bf{\hat{y}},\bf x)=-\text{WMD}(\bf{\hat{y}},\bf x) 
\end{equation}
Compared with other similarity measures such as like cosine similarity, WMD focus more on the cumulative cost of transforming sentence $\bf{\hat{y}}$ into $\bf x$. Besides, WMD considers the entire semantics of the sentence and can effectively handle the synonymic-similarity at the word-level \cite{imat}. The calculation formulate of WMD is as follows:
\begin{equation}
    \text{WMD}({\bf \hat{y}},{\bf x}) = \min_{\mathbf{T}\geqslant 0} \sum_{i,j=1}^{v}(\mathbf{T}_{{ij}}\cdot c(i,j))
\end{equation}
where $v$ is the vocabulary size and $\mathbf{T}_{{ij}}$ is the distance traveled from the word $i$ in $\bf{\hat{y}}$ to the word $j$ in $\bf x$, and its corresponding cost is $c(i,j)$.

\subsubsection{Style Intensity} 
In addition to the semantic similarity, we also expect $\bf{\hat{y}}$ to have strong style intensity towards $\mathcal{Y}$.
Thus, we pre-train Text-CNN-based style classifier \cite{textcnn} on the training corpus and use this classifier to evaluate the style intensity of $\bf{\hat{y}}$, which is estimated by the probability that $\bf{\hat{y}}$ belongs to style $\mathcal{Y}$:
\begin{equation}
    \bf{Sty}(\bf{\hat{y}},\mathcal{Y}) = P_{cls}(\mathcal{Y}|\bf{\hat{y}};\theta_{cls}) \label{eq:style_int}
\end{equation}

Therefore, the final quality score of $\bf{\hat{y}}$ with regard to sentence $\bf x$ and style $\mathcal{Y}$ is calculated as follow:
\begin{equation}
\begin{aligned}
    \bf{CS}(\hat{\bf{y}},\bf{x},\mathcal{Y})=\bf{Con}(\bf{\hat{y}},\bf x)+\bf{Sty}(\bf{\hat{y}},\mathcal{Y})
\end{aligned}
\label{eq:sc}
\end{equation}

Based on the score given by $\bf{CS}$, the refinement mechanism chooses the sentence $\bf \hat{y}$ with the maximal score in $\mathbb{C}_{\bf x}$ as the best sentence $\bf{y}^*$ and paired with $\bf x$ as the pseudo-parallel data for the next epoch of training. 

Overall, the refinement mechanism for $\bf{y}^*$ can be formulated as follow:
\begin{equation}
    \bf{y}^* =\argmax_{\bf{\hat{y}}\in\mathbb{C}_{\bf x}}\,\bf{CS}(\bf{\hat{y}},\bf x,\mathcal{Y}) \label{eq:rf_y}
\end{equation} 
Similarly, $\bf x^*$ can be formulated as follow:
\begin{equation}
    \bf{x}^* =\argmax_{\bf{\hat{x}}\in\mathbb{C}_{\bf y}}\,\bf{CS}(\bf{\hat{x}},\bf y,\mathcal{X}) \label{eq:rf_x}
\end{equation}
%where $n$ is the number of training epochs.
At last, the pseudo-parallel data after refinement becomes $[\bf{x}^*,\bf{y}]$ and $[\bf{y}^*,\bf{x}]$.

\subsection{Style Denoising} \label{sec:sn}
Although the refinement mechanism in our model can highly improve the quality of pseudo-parallel data, the conflict between the goal of the conventional denoising process and the goal of the text style transfer task makes it difficult for the DAE to translate rare style words accurately. In the conventional denoising process, all the information of the original sentences is required to be recovered. However, in the text style transfer task, only the semantic information needs  to be preserved while the style information needs to be converted. As a consequence of this, the DAE trained using the conventional denoising process tends to treat those style words that is rarely appeared in the training set as content words, and mistakenly retains these style words in the model output. 

In this work, we handle the above conflict through a specifically designed style denoising mechanism: First, we use a pre-trained logistic regression classifier \cite{evaluating} to identify a set of style words for each style. Then, in the style transfer training procedure, we randomly replace the input tokens with the identified style words as a type of noise and enforce the DAE model to eliminate such noise in the output sentences.

Specifically, we "\textit{pollute}" $\bf{x}^*$ (or $\bf{y}^*$) by randomly substituting input words in $\bf{x}^*$ (or $\bf{y}^*$) with style words of $\mathcal{X}$ (or $\mathcal{Y}$) by a probability of $\bf{p_{sn}}$, and obtain the corrupted sentences $\bf{x^*_{sn}}$ (or $\bf{y^*_{sn}}$):
\begin{eqnarray}
    \bf{x^*_{sn}} = \bf{Pollute}({\bf x^*},\mathbb{X},p_{sn}) \label{eq:sn_x} \\
% \end{eqnarray}
% and
% \begin{equation}
    \bf{y^*_{sn}} = \bf{Pollute}(\bf{y^*},\mathbb{Y},p_{sn}) \label{eq:sn_y}
\end{eqnarray}
$\mathbb{X}$ and $\mathbb{Y}$ are sets of style words for style $\mathcal{X}$ and $\mathcal{Y}$ respectively. Function $\bf{Pollute}(\bf s,\mathbb{M},\bf{p})$ represents the process of randomly substituting words in $\bf s$ with words in $\mathbb{M}$ with the probability of $\bf{p}$.

Then, in the style transfer process, we expect the model to convert the added style words and transform the corrupted sentence $\bf{x^*_{sn}}$ (or $\bf{y^*_{sn}}$) into the ground-truth sentence $\bf{y}$ (or $\bf x$), by minimizing the following MLE loss:
\begin{eqnarray}
    \mathcal{L}_{\bf{MLE}}=\log{\bf P(\bf x | \bf{y^*_{sn}})}+\log{\bf P(\bf y | \bf{x^*_{sn}})}
\end{eqnarray}

Finally, to verify that the added ``style noise'' has been removed and the output sentence has the desired style, we utilize the same pre-trained style classifier ( same as the one used in Eq.\ref{eq:style_int}) to estimate the style preference of output sentences and expect our model to maximize the probability that output sentences have target-style attributes:
\begin{equation}
    \mathcal{L}_{\bf{style}} =\log{\bf P_{cls}(\mathcal{Y}|\bf{x}_{_\mathcal{Y}})}
    +\log{\bf P_{cls}(\mathcal{X}|\bf{y}_{_\mathcal{X}})} \label{eq:style}
\end{equation}
where $\bf{x}_{_\mathcal{Y}}$ and $\bf{y}_{_\mathcal{X}}$ are the transferred output of the model, with $\bf{x^*_{sn}}$ and $\bf{y^*_{sn}}$ as input, respectively.

There are several advantages to insert style noise into the inputs. Firstly, the added style words can enlarge the style gap between the input ($\bf{x^*}$ and $\bf{y^*}$) and the ground-truth ($\bf{y}$ and $\bf{x}$), and help the model better capture the differences between each style. Secondly, rare style words will appear more frequently in the input sentences, which can help the model better handle the transfer of these rare style words. Moreover, the goal of the style denoising process is consistent with the purpose of the style transfer task, which enforces the model to learn to remove the source style words in the original sentences. In general, the style denoising mechanism makes the model more sensitive to the style words and enhances the ability of the model's ability to convert style words of input sentences.

\subsubsection{Final Training Loss}
Overall, the final training loss used for the complete style transfer process is:
\begin{equation}
    \mathcal{L} = 
    %\lambda_{1}\cdot\mathcal{L}_{\text{SM}} + 
    \lambda\cdot\mathcal{L}_{\bf{MLE}} + (1-\lambda)\cdot\mathcal{L}_{\bf{style}} \label{eq:final_l}
\end{equation}
where $\lambda$ is a tunable parameter that is used to balance the content preservation and style accuracy. 
It is worth mentioning that we do not back-propagate the gradient of $\mathcal{L}$ in the back-translation part ( ${\bf M}_{\mathcal{X} \Leftrightarrow \mathcal{Y}}^{\bf t}$ in the Figure \ref{fig:2}).

\section{Experiments}
\subsection{Datasets}
We verify the effectiveness of our model on two style transfer tasks: sentiment transfer task and formality transfer task. Table \ref{table:data-sta} shows the statistics of datasets used in the two transfer tasks.

\begin{table}[h]
\small
\caption{Statistics of datasets (\textsc{Yelp} and \textsc{Gyafc}).}
\centering
\begin{tabular}{c|cc|cc}
\hline
\multirow{2}{*}{} &
\multicolumn{2}{c|}{\textsc{Yelp}} &
\multicolumn{2}{c}{\textsc{Gyafc}} \\
 & Positive &Negative & Formal & Informal \\
\hline
Train& 266,041  & 177,218  & 51,967    & 51,967 \\
Dev&2,000 & 2,000 & 2,247  & 2,788 \\
Test&500 &500  & 500  & 500  \\
\hline
\end{tabular}
\vspace{-1em}
\label{table:data-sta}.
\end{table}

\begin{table*}[!h]

\caption{Automatic evaluation results of different models on two benchmark datasets.}
\begin{center}
\begin{tabular}{l|cccc|cccc}
\hline
\multirow{2}{*}{} &
\multicolumn{4}{c|}{\textsc{Yelp}} &
\multicolumn{4}{c}{\textsc{Gyafc}} \\
 & Accuracy &BLEU-\textit{r} & \textbf{G2} & \textbf{H2} & Accuracy & BLEU-\textit{r} & \textbf{G2}& \textbf{H2} \\
\hline
% Input Copy&2.5  & 64.3   & 12.7  & 4.8  & 13.4 & 48.2   & 25.4  & 21.0 \\  
% Human Reference &74.8   & 100.0    & 86.5   & 85.6   & 85.7   & 100.0    & 92.6   & 92.3   \\
% \hline
\multicolumn{9}{c}{ \textit{Existing Unsupervised Systems}} \\
\hline
CrossAlign \cite{shen-ca}&73.8  & 16.8   & 35.2  & 27.4  & 69.8  & 3.6   & 15.8  & 6.8  \\
StyleEmbedding \cite{fu-gan}&8.7  & 38.5   & 18.3  & 14.2   & 22.2  & 7.9   & 13.2  & 11.7  \\
% MultiDec \cite{fu-gan}&49.8  & 25.4    & 37.3  & 35.8  & 15.9  & 12.3   & 14.0  & 13.9  \\
% TempBased \cite{li-2018-delete}&78.8  & 42.1    & 59.9  & 57.7  & 50.9  & 35.2  & 42.3  & 41.6  \\
% RetrivOnly \cite{li-2018-delete}&95.0  & 3.0    & 16.7  & 5.7  & \textbf{90.9}  & 0.4   & 6.0  & 0.8  \\
DelRetri \cite{li-2018-delete}&89.2  & 28.7   & 50.6   & 43.5   & 54.6  & 21.2   & 34.0  & 30.5  \\
UnsupervisedMT \cite{zhang-umt}& {96.3}  & 41.8    & 63.4   & 58.3   & 68.9  & 33.4  & 48.0  & 45.0  \\
BackTranslation \cite{back-translation}&81.4   & 37.2    & 55.0   & 51.1   & 26.3   & 44.5    & 34.2   & 33.0  \\
MultiAttribute \cite{lample2019}&81.4  & 39.6  & 56.8  & 53.3  & 26.3  & 44.5    & 34.2  & 33.0   \\
% TransfomerC \cite{style-gan}&90.6  & 48.8  & 66.5  & 63.4  & 34.4  & 43.8   & 38.8  & 38.5  \\
TransfomerGAN \cite{style-gan}&85.4  & 51.4    & 66.2   & 64.1   & 66.4  & 44.6   & 54.4  & 53.4  \\
DualRL \cite{luo19a}&88.6  & 52.0    & 67.9   & 65.5   & 62.6  & 41.9   & 51.2  & 50.2  \\
DataRefinement \cite{imat}&87.7  & 19.8    & 41.7   & 32.4   & 64.1   & { 50.7}  & 57.0   & 56.6   \\
PoiGen \cite{wu-rl} & 86.8   & 54.4   & 68.7   & 66.9  & 11.7  & 44.9  & 22.9  & 18.6  \\
WordRelev \cite{zhou-etal-2020-exploring} & 84.3 & {57.0} & {69.3} & {68.0} & 81.8 & 46.2   & 61.4  & 59.0  \\
StyleIns \cite{ijcai2020-526} & 91.1 & 45.5  & 64.4  & 60.7  & 69.6 & 47.5  & 57.5  & 56.5  \\
% WordRelev \cite{zhou-etal-2020-exploring} & {94.0} & \textbf{60.4} & \textbf{75.4} & \textbf{73.6} & {81.4} & \textbf{47.7} & \textbf{62.3} & \textbf{60.2} \\
\hline
\multicolumn{9}{c}{ \textit{Unsupervised Baseline Systems}} \\
\hline
Input Copy&2.5  & 58.6   & 12.1   & 4.8  & 13.4 & 48.2   & 25.4  & 21.0 \\
Vanilla DAE&73.3 &54.3  &63.1 &62.4 & 29.2 &47.6  &37.3  &36.2 \\
\hline
\multicolumn{9}{c}{\textit{Our Unsupervised DAE Systems}} \\
\hline
%\qquad+ Classifier & 82.3 &52.0 &65.4 &63.7 &81.9 &47.4  &62.3  &60.0 \\
%\qquad+ Data Refinement & 82.1 & \underline{57.5} & 68.7 & 67.6 &81.7 & \underline{\textbf{48.8}} & 63.2  & 61.1 \\
SE-DAE & {85.7}   &{53.9}    & {68.0}   & {66.2}   & {85.5}  & 48.4    & {{64.3}}   & {{61.8}}   \\
\hline
\end{tabular}
\end{center}
\vspace{-1em}
\label{tab:main-exp}
\end{table*}

\noindent \textbf{Sentiment transfer} We test the sentiment transfer  capability of the model on \textbf{\textsc{Yelp}} review dataset, which includes positive and negative reviews about restaurants and businesses. In line with prior works, we use the pre-processed dataset from Luo et al.\cite{luo19a}, which has the same dividing on train, dev and test sets as Li et al.\cite{li-2018-delete}. The five human references of the test set proposed in Jin et al.\cite{imat} are used to evaluate the performance of content preservation.

\noindent \textbf{Formality transfer} The representative dataset \textbf{\textsc{Gyafc}} (\textbf{G}rammarly’s \textbf{Y}ahoo \textbf{A}nswers \textbf{F}ormality \textbf{C}orpus) is used to estimate the formality transfer capacity of the model \cite{rao-tetreault-2018-dear}. Following Luo et al.\cite{luo19a}, we choose the \textsc{Family And Relationships} domain and treat it as a non-parallel dataset.

\begin{table*}[!h]
\centering
\caption{Human evaluation results of different models on two benchmark datasets.}
\begin{tabular}{l|ccc|ccc}
\hline
\multirow{2}{*}{} &
\multicolumn{3}{c|}{\textsc{Yelp}} &
\multicolumn{3}{c}{\textsc{Gyafc}} \\
 & Style & Content & Fluency & Style & Content & Fluency   \\
\hline
%MultiAttri \cite{lample2019} & 
%3.53 & 4.04 & 4.13 & 3.90   & 
%2.38 & \textbf{4.66}  & \textbf{4.50}  & 3.85   \\
TransfomerGAN \cite{style-gan} & 
3.65 \begin{tiny}${\pm \bf 1.05}$\end{tiny}& 
4.33 \begin{tiny}${\pm \bf 0.70}$\end{tiny}& 
4.06 \begin{tiny}${\pm \bf 0.83}$\end{tiny}& 
%4.01 \;\begin{tiny}${\pm 1.29}$\end{tiny}& 
2.75 \begin{tiny}${\pm \bf 0.85}$\end{tiny}& 
3.85 \begin{tiny}${\pm \bf 0.89}$\end{tiny}& 
3.67 \begin{tiny}${\pm \bf 0.97}$\end{tiny} \\
%3.42 \;\begin{tiny}${\pm 1.29}$\end{tiny} 

DualRL \cite{luo19a} & 
3.90 \begin{tiny}${\pm \bf 0.93}$\end{tiny}& 
4.30 \begin{tiny}${\pm \bf 0.88}$\end{tiny}& 
4.21 \begin{tiny}${\pm \bf 0.83}$\end{tiny}& 
%4.13 \;\begin{tiny}${\pm 1.29}$\end{tiny}& 
2.57 \begin{tiny}${\pm \bf 0.77}$\end{tiny}& 
3.61 \begin{tiny}${\pm \bf 1.23}$\end{tiny}& 
3.73 \begin{tiny}${\pm \bf 0.95}$\end{tiny} \\
%3.30 \;\begin{tiny}${\pm 1.29}$\end{tiny}  

PoiGen \cite{wu-rl} & 
3.83 \begin{tiny}${\pm \bf 1.13}$\end{tiny}& 
4.46 \begin{tiny}${\pm \bf 0.74}$\end{tiny}& 
4.37 \begin{tiny}${\pm \bf 0.72}$\end{tiny}& 
%4.22 \;\begin{tiny}${\pm 1.29}$\end{tiny}& 
1.56 \begin{tiny}${\pm \bf 0.74}$\end{tiny}& 
2.77 \begin{tiny}${\pm \bf 0.52}$\end{tiny}& 
2.88 \begin{tiny}${\pm \bf 0.71}$\end{tiny} \\
%2.41 \;\begin{tiny}${\pm 1.29}$\end{tiny}  

WordRelev \cite{zhou-etal-2020-exploring} & 
{4.03} \begin{tiny}${\pm \bf 0.95}$\end{tiny}& 
{4.63} \begin{tiny}${\pm \bf 0.55}$\end{tiny}& 
{4.50} \begin{tiny}${\pm \bf 0.56}$\end{tiny}& 
%\textbf{4.39} \;\begin{tiny}${\pm 1.29}$\end{tiny}& 
3.47 \begin{tiny}${\pm \bf 0.86}$\end{tiny}& 
4.35 \begin{tiny}${\pm \bf 0.72}$\end{tiny}& 
{4.37} \begin{tiny}${\pm \bf 0.91}$\end{tiny} \\
%4.05  \;\begin{tiny}${\pm 1.29}$\end{tiny}  

StyleIns \cite{ijcai2020-526} & 
3.82 \begin{tiny}${\pm \bf 1.02}$\end{tiny}& 
4.12 \begin{tiny}${\pm \bf 0.93}$\end{tiny}& 
4.11 \begin{tiny}${\pm \bf 0.87}$\end{tiny}& 
%4.02 \;\begin{tiny}${\pm 1.29}$\end{tiny}& 
3.27 \begin{tiny}${\pm \bf 1.03}$\end{tiny}& 
4.33 \begin{tiny}${\pm \bf 0.66}$\end{tiny}& 
4.32 \begin{tiny}${\pm \bf 0.77}$\end{tiny} \\
%3.98 \;\begin{tiny}${\pm 1.29}$\end{tiny}  
\hline
SE-DAE &
{3.88} \begin{tiny}${\pm \bf 0.98}$\end{tiny}& 
{4.48} \begin{tiny}${\pm \bf 0.63}$\end{tiny}& 
{4.38} \begin{tiny}${\pm \bf 0.69}$\end{tiny}& 
%{4.25} \;\begin{tiny}${\pm 1.29}$\end{tiny}& 
{3.55} \begin{tiny}${\pm \bf 1.01}$\end{tiny}& 
{4.38} \begin{tiny}${\pm \bf 0.68}$\end{tiny}& 
{4.36} \begin{tiny}${\pm \bf 0.75}$\end{tiny}\\
%\textbf{4.10} \;\begin{tiny}${\pm 1.29}$\end{tiny} 
\hline
\end{tabular}
\vspace{-1em}
\label{tab:human-exp}
\end{table*}

\begin{table*}[!h]
\centering
\caption{Ablation study of different components used in SE-DAE.} 
\begin{tabular}{l|cccccc|cccccc}
\hline
\multirow{2}{*}{} &
\multicolumn{6}{c|}{\textsc{Yelp}} &
\multicolumn{6}{c}{\textsc{Gyafc}} \\
 & ACC & BLEU-\textit{r}  & \textbf{G2} &$\Delta$ &\textbf{H2} &$\Delta$ & ACC & BLEU-\textit{r} & \textbf{G2} &$\Delta$ &\textbf{H2} &$\Delta$ \\
\hline
Vanilla DAE&73.3 &{51.9}  &63.1  &\textbf{0.0} & 62.4  &\textbf{0.0} & 29.2 &47.6  &37.3 &\textbf{0.0} &36.2 &\textbf{0.0}  \\
\qquad+ Classifier & 82.3 &48.7  &63.3  &{\color{forestgreen}\bf{+0.2}}  &61.2  &{\color{falured}\bf{-1.2}} &81.9 &47.4 &62.3  &{\color{forestgreen}\bf{+25.0}}  &60.0  &{\color{forestgreen}\bf{+23.8}} \\
\qquad+ Refinement & 82.1 &53.6  &66.4 &{\color{forestgreen}\bf{+3.3}} &64.9 &{\color{forestgreen}\bf{+2.5}}  &81.7 &{48.8} &63.2 &{\color{forestgreen}\bf{+25.9}} &61.1  &{\color{forestgreen}\bf{+24.9}} \\
\qquad+ Style Noise&{85.7}   &53.9 &68.0  &{\color{forestgreen}\bf{+4.9}}  &66.2    & {\color{forestgreen}\bf{+3.8}}   & {{85.5}}   &48.4 &64.3 &{\color{forestgreen}\bf{+27.0}}    & 61.8  & {\color{forestgreen}\bf{+25.6}}   \\
\hline
\end{tabular}
\vspace{-1em}
\label{tab:ab}
\end{table*}

\subsection{Baselines}
We mainly compare our model with the latest SOTA style-transfer models, which have opensourced their code : TransfomerGAN \cite{style-gan}, DualRL \cite{luo19a}, PoiGen \cite{wu-rl}, StyleIns \cite{ijcai2020-526} and WordRelev \cite{zhou-etal-2020-exploring}. Besides, we also include earlier transfer systems for comprehensive comparison, such as UnsuperMT \cite{zhang-umt} and BackTranslation \cite{back-translation}. 

\subsection{Training Details}
In this work, we implement both encoder and decoder of our model on two-layer bidirectional LSTM with attention mechanism \cite{lstm-attention}.We set dimensions of all embedding to 512 and train them from scratch. The Adam optimizer is used in our model with a learning rate of $10^{-4}$ during the whole training process. The batch size used in our model is set to 32 and the $\bf p_{sn}$ is set to 0.2. Besides, following Mir et al.\cite{evaluating}, we obtain style words of different styles via the logistic regression model with a high threshold of 0.9 to reduce noise. We use the same pre-train style classifier in Eq.\ref{eq:sc} and \ref{eq:style} which is implemented on Text-CNN \cite{textcnn} and achieves accuracy of 97\% and 90\% on \textsc{Yelp} and \textsc{Gyafc} respectively. We keep $\lambda=0.3$ during the whole training process. 
% Finally, we train our model on one TITAN Xp GPU for nearly 20 hours (and nearly 60 epochs) until it converges and the total parameters of our model are nearly 30M.

\subsection{Metrics}
\subsubsection{Automatic Metrics}
\noindent \textbf{\em{Style Accuracy}} To evaluate the style accuracy of transferred sentences, we train two TextCNN-based style classifiers (different from classifiers used in the training process), which achieves  97\% and 89\% accuracy on \textsc{Yelp} and \textsc{Gyafc} respectively. To distinguish from the style classifier used in the training process, these two classifiers are implemented by Tensorflow.

\noindent \textbf{\em Content Preservation} In this work, the BLEU score between the inference of the style-transfer model and four human references is used to evaluate the content preservation performance of the model.

\noindent \textbf{\em Overall Performance} We calculate the geometric mean (G2) and harmonic mean (H2) of style accuracy and BLEU score to evaluate the overall performance of each model, which is also used in Luo et al.\cite{luo19a},

\subsubsection{Human Metrics}
We randomly select 400 sentences of different styles (i.e., 100 positive and 100 negative; 100 informal and 100 formal.) and hire five human judges with strong linguistic backgrounds (having learned English for more than fifteen years) to evaluate the quality of transferred sentences. Without any information about which model the sentence is generated by, the five human judges are asked to rate the output sentences of different style-transfer models on a Likert scale of 1 to 5 to evaluate the target style accuracy, content preservation and sentence fluency (Fluency).

\begin{figure*}
\centering
\includegraphics[width=1.85\columnwidth]{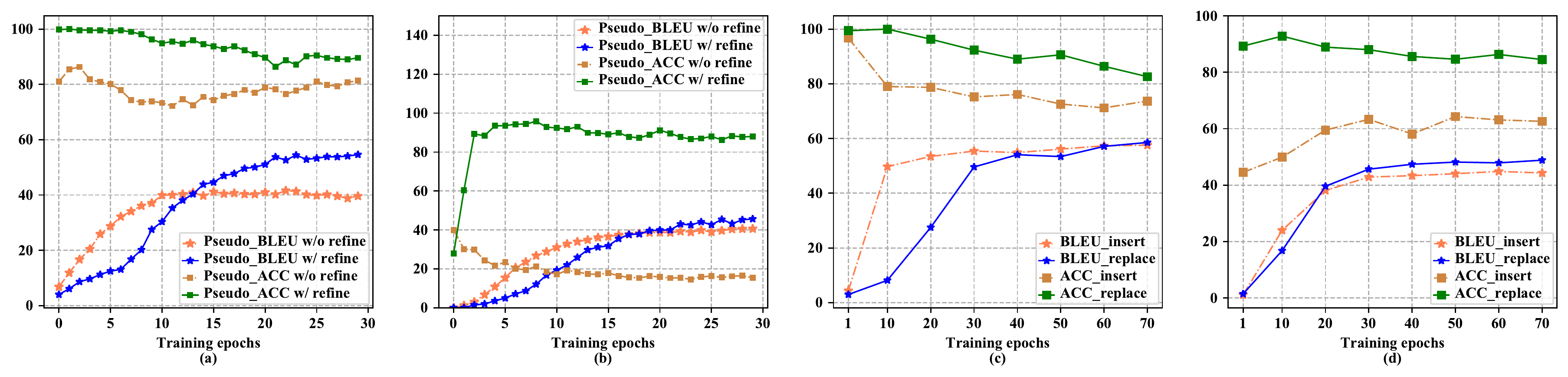}
\caption{\textit{(a)}: Evolution of BLEU score and style accuracy of the pseudo-parallel data with or without refinement on \textsc{Yelp}. \textit{(b)}: Evolution of BLEU score and style accuracy of the pseudo-parallel data with or without refinement on \textsc{Gyafc}. \textit{(c)}: Evolution of BLEU score and style-accuracy on \textsc{Yelp} when using different ways to add style words. \textit{(d)}: Evolution of BLEU score and style-accuracy on \textsc{Gyafc} when using different ways to add style words. }
\vspace{-1em}
\label{fig:4} 
\end{figure*}

\begin{figure*}
\centering
\includegraphics[width=2.0\columnwidth]{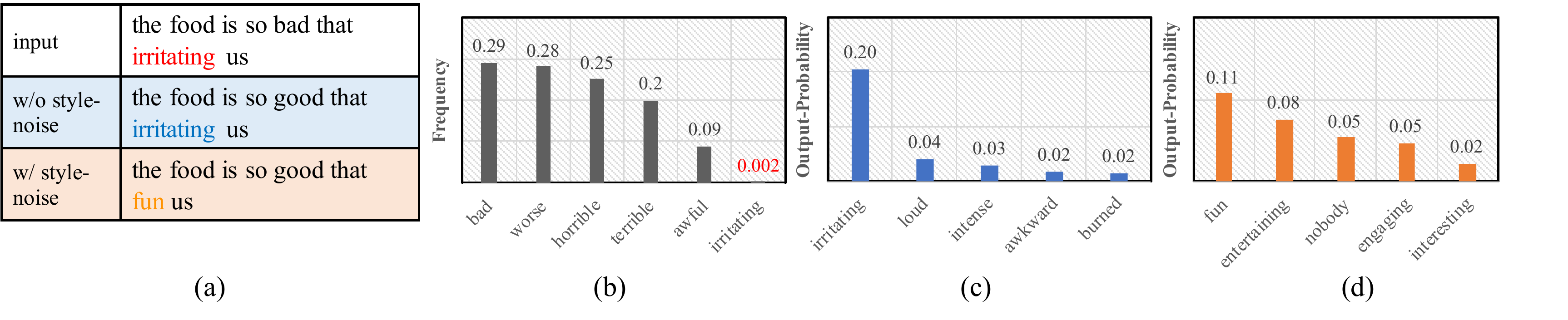}
\caption{\textit{(a)}: An example on \textsc{Yelp}, style-noise has an influence on transferring negative style words. \textit{(b)}: Appearance frequency of different negative words on \textsc{Yelp}. \textit{(c)}: Top-5 output words of the model trained \textbf{without} style-noise when transferring the rare negative word ``irritating''.  \textit{(d)}: Top-5 output words of the model trained \textbf{with} style-noise when transferring the rare negative word ``irritating''.} 
\vspace{-1em}
\label{fig:5} 
\end{figure*}

\section{Evaluation Results}
Table \ref{tab:main-exp} presents the results of \textbf{automatic evaluation} of different models (\textbf{G2} and \textbf{H2} representing the geometric mean and harmonic mean of \textit{accuracy} and \textit{BLEU-\textit{r}} respectively. ). In \textsc{Yelp}, our model outperforms other style-transfer models except for WordRelev and PoiGen. The reason is that both WordRelev and PoiGen utilize the word-level style information through a multi-stage training process, but our model is only uses the sentence-level style information of pesudo-parallel data for training. In contrast, our model achieves the best performance on \textsc{Gyafc}. A possible explanation is that there are more abbreviations and spell errors in sentences of \textsc{Gyafc} than that in \textsc{Yelp}, so models are required to be more robust to noise when transferring the style of sentences of \textsc{Gyafc}. Thus, compared with other models, our model that based on denoising process is more robust to the noise and can better transfer the sentences of \textsc{Gyafc}. 
The results of \textbf{human evaluation} are shown in Table \ref{tab:human-exp}. Similar to the results of the automatic evaluation, our model surpasses most previous models except for the WordRelev \cite{zhou-etal-2020-exploring} on \textsc{Yelp}.When \textsc{Gyafc} is involved, our model can achieve similar performance to WordRelev and has higher style accuracy and content score. Overall, the evaluation results show the effectiveness of our model and a competitive advantage over those powerful but complex models.

\section{Performance Analysis}
\subsection{Ablation Study}
To get a better understanding of roles played by different components in our model, we conduct an ablation studies for each components in our model by both \textsc{Yelp} and \textsc{Gyafc}, and present the results in Table \ref{tab:ab}. The results show that the refinement mechanism highly improves BLEU scores on more than 5 points on \textsc{Yelp} and by more than 1 point on \textsc{Gyafc}. Meanwhile, the style denoising mechanism improves the style accuracy on both two datasets by over 3 percentage. In addition, the results of the ablation studies also show that the style classifier plays a crucial role in balancing the style intensity and content preservation. Without the style classifier, the vanilla DAE model will generate sentences that can successfully preserves content information but fails in transferring styles.

\subsection{Refine Pesudo-parallel Data}
In this section, we investigate the influence of the data refinement mechanism by visualizing the evolution curves of the BLEU and style accuracy of the pseudo-parallel data in each epoch, as shown in Figure \ref{fig:4} (a) and (b). The result indicates the effectiveness of data refinement mechanism in finding the sentences that both preserving the semantics and transferring the style. Moreover, the result also shows that without the data refinement mechanism, the pseudo-parallel data suffers from more severe data-quality problem on \textsc{Gyafc} than on \textsc{Yelp}. A possible explanation is that the style of sentiment is more distinguishable than the style of formality and transferring sentiment is much easier than transferring formality.

\begin{table*}[t]
\centering
\caption{Example outputs of different transfer models on sentiment and formality transfer tasks. }
\begin{tabular}{l|l|l}
\hline
 & \textbf{From {\color{red} Negative} to {\color{blue} Positive}} & \textbf{From {\color{blue} Positive} to {\color{red} Negative}}  \\
\hline
\textbf{Source} &  {\textbf{if i could give {\color{red}less} stars , i would .}} & {\textbf{thank you ladies for being {\color{blue}awesome} !}}\\
\hline
%\text{MultiAttri} & {if i could give less stars , i would .} & {no ladies are \text{waiting for these days} !}\\
\text{TransfomerGAN} &  {if i can give \textbf{\color{blue}fun} place , i eat .} & {not you ladies for being \textbf{\color{blue}awesome} !}\\
\text{DualRL} & {if i could give it a shot .} & {\text{why are} the ladies for being \textbf{\color{blue}awesome} ?}\\
\text{PoiGen} & {great if i could give  \textbf{\color{red}less} stars , i would .} & {no thank you for being \textbf{\color{blue}awesome} !}\\
\text{StyleIns} & {thank you for and \textbf{\color{blue}outstanding} service , and service .} & {are i asked for being \textbf{\color{red}disgusting} .}\\
\text{WordRelev} & {if i could give \textbf{\color{blue}extra} stars , i would .} & {thank you ladies for being \textbf{\color{red}horrible} !}\\
\hline
\text{SE-DAE} & {if i can give \textbf{\color{blue}more} stars , i recommend !} & {\text{shame on} ladies for being \textbf{\color{red}nasty} !} \\
\hline
% \text{Human} & {i would give you more stars if i could} & {no thank you ladies for being awful ! } \\
% \hline
\multicolumn{3}{c}{}\\
\hline
 & \textbf{From {\color{ao} Fomal} to {\color{desert} Informal}} & \textbf{From {\color{desert} Informal} to {\color{ao} Fomal}} \\
\hline
\textbf{Source} & \makecell[l]{ \textbf{you should play sports with your male friends .}} & {\textbf{{\color{desert}r u} talking about {\color{desert}ur} avatar ?}}\\
\hline
%\text{MultiAttri} & \makecell[l]{you should play sports with your male friends .} & {\text{r u }talking about \text{ur} avatar ?}\\
\text{TransfomerGAN} & \makecell[l]{and \textbf{\color{desert}just} play sports with your \textbf{\color{desert}9 :} for just !} & {\text{exist} \textbf{\color{ao}you} talking about !}\\
\text{DualRL} & \makecell[l]{you should play sports with your male friends} & {\textbf{\color{ao}it is} talking about \text{avatar} avatar ?}\\
\text{PoiGen} & \makecell[l]{you play with your male friends .} & {\textbf{\color{desert}u} talking about avatar ?}\\
\text{StyleIns} & \makecell[l]{and if play sports with your male friends .} & {\textbf{\color{desert}r u} talking about \textbf{\color{desert}ur} avatar ?}\\
\text{WordRelev} & \makecell[l]{you should play sports with your male friends bout this .} & {\textbf{\color{ao}it is} talking about \textbf{\color{ao}your} avatar ?}\\
\hline
\text{SE-DAE} & \makecell[l]{\textbf{\color{desert}u just} play sports with \textbf{\color{desert}ur} male friends} & {\textbf{\color{ao}are you} talking about \textbf{\color{ao}your} avatar ?}\\
\hline
\end{tabular}
\vspace{-1em}
\label{tab:cs_exp}
\end{table*}

\subsection{Which is Better: Replace Or Insert?}
In this work, we explore two ways to add style words in the style denoising mechanism: (1) randomly \textit{replacing} input words with style words and (2) randomly \textit{inserting} style words into input sentences. The result (see Figure \ref{fig:4} (c) and (d)) indicates it is much better to add style words through replacement than insertion in this task. One possible explanation is that the insertion does not corrupt the original information of sentences, and the model trained by insertion tends to filter words (i.e., focus on filtering out style words). On the contrary, the replacement will break the information of both content and style of the input sentences. As a result, the model trained with replacement needs to recover the content and style information given partial information as input (i.e., focus on the reconstruction of both the semantic and the style).

\subsection{Transfer Style words} \label{sec:c_study}
To further investigate the influence of the style denoising mechanism on transferring the rare style words, we sample a sentence that contains the rare negative word \textit{irritating}. As shown in Figure \ref{fig:5} (b), \textit{irritating} rarely appears in \textsc{Yelp} with a low frequency of 0.002 while \textit{bad} appears most frequently with a probability of 0.29. When transferring \textit{irritating}, the model without style denoising mechanism tends to treat this style word as a content word and gives the highest output probability to it (Figure \ref{fig:5} (c)). In contrast, as the style denoising mechanism increases the appearance frequency of rare style words, the model with style denoising mechanism provides higher output probability to words with the opposite style and outputs the positive word ``fun'' (Figure \ref{fig:5} (d)).

\subsection{Case Study} \label{sec:c_study}
To scrutinize the performance of different transfer models, we sample several sentences with multiple style words or complex syntactic structures and then compare the corresponding transferred sentences of our model with those of other state-of-the-art representative systems.From the result shown in Table \ref{tab:cs_exp}, we observe that: 1) Our model can precisely convert each style words of source sentences 2) In contrast to the incoherent sentences of other models, the sentences generated by our model are more understandable.

\section{Related Work}
Most of the existing models achieve the style transfer of texts from two perspectives: model-oriented and data-oriented.

Model-oriented methods handle the text style transfer problem by designing complicated neural networks, which may incorporate multiple steps in the training procedure. Shen et al.\cite{shen-ca} design a powerful model that can align two different style domains by adversarial learning. Fu et al.\cite{fu-gan} applied an adversarial loss on the encoder to separate style and content information, while Dai et al.\cite{style-gan} uses cyle-GAN to achieve style transfer without disentangling the style and content. Meanwhile, Xu et al.\cite{cycle-rl} proposed a transfer model based on cycle reinforcement-learning, and Luo et al.\cite{luo19a,luo19b} solved the text style transfer problem by dual reinforcement learning. Both Xu et al.\cite{cycle-rl} and Luo et al.\cite{luo19a,luo19b} contain a two-step training procedure. Gong et al.\cite{gong-rl} designed another RL-based model, which considers the style, semantic and fluency rewards of the transferred sentence simultaneously. Wu et al.\cite{wu-rl} proposed a multi-step model by hierarchical reinforcement learning. Liu et al.\cite{Liu-vae} proposed a VAE model that transfers the style of the sentence by modifying its latent representation in the continuous space.

Data-oriented methods prefer to retrieve more word-level or sentences-level information from training corpus for transfer models. Li et al.\cite{li-2018-delete} proposed a retrieval-based transfer model that could retrieve similar sentences from target-style corpus to help the transfer of source-style sentences. Sudhakar et al.\cite{Sudhakar} improved the retrieval process of Li et al.\cite{li-2018-delete} with a more complicated scoring function and proposed another method of use style-embedding as style input when there are no similar sentences in the target style corpus. Lample et al.\cite{lample2019} on-the-fly generated pseudo-parallel data and adopted the unsupervised machine translation model \cite{lample-etal-2018-phrase} as their transfer model. Jin et al.\cite{imat} iteratively back-translated and matched unpaired corpus to obtain high-quality pseudo-parallel data to train their transfer model, but they only considered content similarity when paired two sentences. Wu et al.\cite{wu2019} propose a two-step model that first masks style words with a pre-trained identifier, and then train the model to infill the masked words with words of the target style. Zhou et al.\cite{zhou-etal-2020-exploring} propose a two-stage model that calculates the word-level style relevance in the first stage and then utilizes the learned style relevance in the second stage.

\section{Conclusion}
In this work, we propose the Style-Enhanced Denoising Auto-Encoder (SE-DAE) with a novel style-denoising mechanism, which is more compatible with style transfer tasks. To preserve the simplicity of original vanilla DAE, we train SE-DAE on the high-quality pseudo-parallel data generated by another novel data-refinement mechanism.
Compared with previously proposed transfer systems that combine the vanilla DAE with complicated unsupervised models, our model achieves competitive performance to them while preserving the simplicity of the DAE in both the network structure and training procedure.
The evaluation results on two benchmark style datasets indicate the effectiveness of SE-DAE in multiple text style transfer tasks.

\bibliography{references}

\begin{thebibliography}{10}
\providecommand{\url}[1]{#1}
\csname url@samestyle\endcsname
\providecommand{\newblock}{\relax}
\providecommand{\bibinfo}[2]{#2}
\providecommand{\BIBentrySTDinterwordspacing}{\spaceskip=0pt\relax}
\providecommand{\BIBentryALTinterwordstretchfactor}{4}
\providecommand{\BIBentryALTinterwordspacing}{\spaceskip=\fontdimen2\font plus
\BIBentryALTinterwordstretchfactor\fontdimen3\font minus
  \fontdimen4\font\relax}
\providecommand{\BIBforeignlanguage}[2]{{%
\expandafter\ifx\csname l@#1\endcsname\relax
\typeout{** WARNING: IEEEtran.bst: No hyphenation pattern has been}%
\typeout{** loaded for the language `#1'. Using the pattern for}%
\typeout{** the default language instead.}%
\else
\language=\csname l@#1\endcsname
\fi
#2}}
\providecommand{\BIBdecl}{\relax}
\BIBdecl

\bibitem{cycle-rl}
J.~Xu, X.~Sun, Q.~Zeng, X.~Ren, X.~Zhang, H.~Wang, and W.~Li, ``Unpaired
  sentiment-to-sentiment translation: A cycled reinforcement learning
  approach,'' in \emph{Proceedings of the 56th Annual Meeting of the
  Association for Computational Linguistics}, 2018, pp. 979--988.

\bibitem{wu-rl}
C.~Wu, X.~Ren, F.~Luo, and X.~Sun, ``A hierarchical reinforced sequence
  operation method for unsupervised text style transfer,'' in \emph{Proceedings
  of the 57th Conference of the Association for Computational Linguistics,
  {ACL}}, 2019, pp. 4873--4883.

\bibitem{luo19a}
F.~Luo, P.~Li, J.~Zhou, P.~Yang, B.~Chang, Z.~Sui, and X.~Sun, ``A dual
  reinforcement learning framework for unsupervised text style transfer,'' in
  \emph{Proceedings of the 28th International Joint Conference on Artificial
  Intelligence, {IJCAI} 2019}, 2019a, pp. 5116--5122.

\bibitem{luo19b}
F.~Luo, P.~Li, P.~Yang, J.~Zhou, Y.~Tan, B.~Chang, Z.~Sui, and X.~Sun,
  ``Towards fine-grained text sentiment transfer,'' in \emph{Proceedings of the
  57th Annual Meeting of the Association for Computational Linguistics, {ACL}
  2019}, 2019b, pp. 2013--2022.

\bibitem{gong-rl}
H.~Gong, S.~Bhat, L.~Wu, J.~Xiong, and W.-m. Hwu, ``Reinforcement learning
  based text style transfer without parallel training corpus,'' in
  \emph{Proceedings of the 2019 Conference of the North {A}merican Chapter of
  the Association for Computational Linguistics: Human Language}, 2019, pp.
  3168--3180.

\bibitem{hu-gan}
Z.~Hu, Z.~Yang, X.~Liang, R.~Salakhutdinov, and E.~P. Xing, ``Toward controlled
  generation of text,'' in \emph{Proceedings of the 34th International
  Conference on Machine Learning}, 2017, pp. 1587--1596.

\bibitem{fu-gan}
Z.~Fu, X.~Tan, N.~Peng, D.~Zhao, and R.~Yan, ``Style transfer in text:
  Exploration and evaluation,'' \emph{Association for the Advancement of
  Artiﬁcial Intelligence}, 2018.

\bibitem{Lajanugen-gan}
L.~Logeswaran, H.~Lee, and S.~Bengio, ``Content preserving text generation with
  attribute controls,'' in \emph{32nd Conference on Neural Information
  Processing Systems (NeurIPS 2018)}, 2018, pp. 5103--5113.

\bibitem{yang-gan}
Z.~Yang, Z.~Hu, C.~Dyer, E.~P. Xing, and T.~Berg-Kirkpatrick, ``Unsupervised
  text style transfer using language models as discriminators,'' in
  \emph{Advances in Neural Information Processing Systems 31}, S.~Bengio,
  H.~Wallach, H.~Larochelle, K.~Grauman, N.~Cesa-Bianchi, and R.~Garnett, Eds.,
  2018, pp. 7287--7298.

\bibitem{lai-gan}
C.-T. Lai, Y.-T. Hong, H.-Y. Chen, C.-J. Lu, and S.-D. Lin, ``Multiple text
  style transfer by using word-level conditional generative adversarial network
  with two-phase training,'' in \emph{Proceedings of the 2019 Conference on
  Empirical Methods in Natural Language Processing and the 9th International
  Joint Conference on Natural Language Processing (EMNLP-IJCNLP)}, 2019, pp.
  3579--3584.

\bibitem{style-gan}
N.~Dai, J.~Liang, X.~Qiu, and X.~Huang, ``Style transformer: Unpaired text
  style transfer without disentangled latent representation,''
  \emph{Proceedings of the 57th Annual Meeting of the Association for
  Computational Linguistics}, pp. 5997--6007, 2019.

\bibitem{tikhonov-etal-2019-style}
A.~Tikhonov, V.~Shibaev, A.~Nagaev, A.~Nugmanova, and I.~P. Yamshchikov,
  ``Style transfer for texts: Retrain, report errors, compare with rewrites,''
  in \emph{Proceedings of the 2019 Conference on Empirical Methods in Natural
  Language Processing and the 9th International Joint Conference on Natural
  Language Processing (EMNLP-IJCNLP)}, 2019, pp. 3936--3945.

\bibitem{lample2019}
G.~Lample, S.~Subramanian, E.~Smith, L.~Denoyer, M.~Ranzato, and Y.-L. Boureau,
  ``Multiple-attribute text rewriting,'' in \emph{International Conference on
  Learning Representations}, 2019.

\bibitem{ub-gan}
H.~Ham, T.~J. Jun, and D.~Kim, ``Unbalanced gans: Pre-training the generator of
  generative adversarial network using variational autoencoder,'' in
  \emph{ArXiv}, 2020.

\bibitem{li-2018-delete}
J.~Li, R.~Jia, H.~He, and P.~Liang, ``Delete, retrieve, generate: a simple
  approach to sentiment and style transfer,'' in \emph{Proceedings of the 2018
  Conference of the North {A}merican Chapter of the Association for
  Computational Linguistics: Human Language Technologies, Volume 1 (Long
  Papers)}, 2018, pp. 1865--1874.

\bibitem{rao-tetreault-2018-dear}
S.~Rao and J.~Tetreault, ``Dear sir or madam, may {I} introduce the {GYAFC}
  dataset: Corpus, benchmarks and metrics for formality style transfer,'' in
  \emph{Proceedings of the 2018 Conference of the North {A}merican Chapter of
  the Association for Computational Linguistics: Human Language Technologies,
  Volume 1 (Long Papers)}, 2018, pp. 129--140.

\bibitem{dae}
P.~Vincent, H.~Larochelle, Y.~Bengio, and P.-A. Manzagol, ``Extracting and
  composing robust features with denoising autoencoders,'' pp. 1096--1103,
  2008.

\bibitem{lample-etal-2018-phrase}
G.~Lample, M.~Ott, A.~Conneau, L.~Denoyer, and M.~Ranzato, ``Phrase-based {\&}
  neural unsupervised machine translation,'' in \emph{Proceedings of the 2018
  Conference on Empirical Methods in Natural Language Processing}, 2018, pp.
  5039--5049.

\bibitem{wmd}
M.~J. Kusner, Y.~Sun, N.~I. Kolkin, and K.~Q. Weinberger, ``From word
  embeddings to document distances,'' in \emph{Proceedings of the 32nd
  International Conference on International Conference on Machine Learning},
  2015, p. 957–966.

\bibitem{imat}
Z.~Jin, D.~Jin, J.~Mueller, N.~Matthews, and E.~Santus, ``{IM}a{T}:
  Unsupervised text attribute transfer via iterative matching and
  translation,'' in \emph{Proceedings of the 2019 Conference on Empirical
  Methods in Natural Language Processing and the 9th International Joint
  Conference on Natural Language Processing (EMNLP-IJCNLP)}, 2019, pp.
  3097--3109.

\bibitem{textcnn}
Y.~Kim, ``Convolutional neural networks for sentence classification,'' in
  \emph{Proceedings of the 2014 Conference on Empirical Methods in Natural
  Language Processing ({EMNLP})}, 2014, pp. 1746--1751.

\bibitem{evaluating}
R.~Mir, B.~Felbo, N.~Obradovich, and I.~Rahwan, ``Evaluating style transfer for
  text,'' in \emph{Proceedings of the 2019 Conference of the North {A}merican
  Chapter of the Association for Computational Linguistics: Human Language
  Technologies, Volume 1 (Long and Short Papers)}, 2019, pp. 495--504.

\bibitem{shen-ca}
T.~Shen, T.~Lei, R.~Barzilay, and T.~S. Jaakkola, ``Style transfer from
  non-parallel text by cross-alignment,'' pp. 6830--6841, 2017.

\bibitem{zhang-umt}
Z.~Zhang, S.~Ren, S.~Liu, J.~Wang, P.~Chen, M.~Li, M.~Zhou, and E.~Chen,
  ``Style transfer as unsupervised machine translation,'' \emph{CoRR}, 2018.

\bibitem{back-translation}
S.~Prabhumoye, Y.~Tsvetkov, R.~Salakhutdinov, and A.~W. Black, ``Style transfer
  through back-translation,'' in \emph{Proceedings of the 56th Annual Meeting
  of the Association for Computational Linguistics (Volume 1: Long Papers)},
  2018, pp. 866--876.

\bibitem{zhou-etal-2020-exploring}
C.~Zhou, L.~Chen, J.~Liu, X.~Xiao, J.~Su, S.~Guo, and H.~Wu, ``Exploring
  contextual word-level style relevance for unsupervised style transfer,'' in
  \emph{Proceedings of the 58th Annual Meeting of the Association for
  Computational Linguistics}, 2020, pp. 7135--7144.

\bibitem{ijcai2020-526}
X.~Yi, Z.~Liu, W.~Li, and M.~Sun, ``Text style transfer via learning style
  instance supported latent space,'' in \emph{Proceedings of the Twenty-Ninth
  International Joint Conference on Artificial Intelligence, {IJCAI-20}}, 2020,
  pp. 3801--3807.

\bibitem{lstm-attention}
D.~Bahdanau, K.~Cho, and Y.~Bengio., ``Neural machine translation by jointly
  learning to align and translate.'' in \emph{In Proceedings of ICLR}, 2015.

\bibitem{Liu-vae}
D.~Liu, J.~Fu, Y.~Zhang, C.~Pal, and J.~Lv, ``Revision in continuous space:
  Unsupervised text style transfer without adversarial learning,'' in
  \emph{AAAI}, 2020.

\bibitem{Sudhakar}
A.~Sudhakar, B.~Upadhyay, and A.~Maheswaran, ``Transforming delete, retrieve,
  generate approach for controlled text style transfer,'' in \emph{Proceedings
  of the 2019 Conference on Empirical Methods in Natural Language Processing
  and the 9th International Joint Conference on Natural Language Processing
  (EMNLP-IJCNLP)}, 2019, pp. 3269--3279.

\bibitem{wu2019}
X.~Wu, T.~Zhang, L.~Zang, J.~Han, and S.~Hu, ``Mask and infill: Applying masked
  language model for sentiment transfer,'' in \emph{Proceedings of the
  Twenty-Eighth International Joint Conference on Artificial Intelligence,
  {IJCAI-19}}, 2019, pp. 5271--5277.

\end{thebibliography}
\bibliographystyle{IEEEtran}

\end{document}